\documentclass[11pt,a4paper]{article}
\usepackage[hyperref]{acl2017}
\usepackage{times}
\usepackage{latexsym}
\usepackage{url}

\usepackage{amsmath}
\usepackage{multirow}
\usepackage{graphicx}
\usepackage{bm}
\usepackage{amssymb}
\usepackage{algorithm2e}
\usepackage{verbatim}
\usepackage{array}
\usepackage{tikz}
\usepackage{xr}

\aclfinalcopy 

\title{Robust Incremental Neural Semantic Graph Parsing}

\author{Jan Buys$^1$ and Phil Blunsom$^{1,2}$ \\
$^1$Department of Computer Science, University of Oxford \qquad
$^2$DeepMind \\
\texttt{\{jan.buys,phil.blunsom\}@cs.ox.ac.uk} \\
}

\date{}

\begin{document}

\maketitle

\begin{abstract}

Parsing sentences to linguistically-expressive 
semantic representations is a key goal of Natural Language Processing.
Yet statistical parsing has focussed almost exclusively on bilexical 
dependencies or domain-specific logical forms.
We propose a neural encoder-decoder transition-based parser which is 
the first full-coverage semantic graph parser for 
Minimal Recursion Semantics (MRS).
The model architecture uses stack-based embedding features, predicting
graphs jointly with 
unlexicalized predicates and their token alignments.
Our parser is more accurate than attention-based
baselines on MRS, and on an additional Abstract Meaning Representation (AMR) benchmark, and
GPU batch processing makes it an order of magnitude faster than a 
high-precision grammar-based parser.
Further, the $86.69\%$ Smatch score of our MRS parser is higher than the upper-bound 
on AMR parsing, making MRS an attractive choice as a semantic
  representation.\footnote{Code, models and data preparation scripts are available at
\url{https://github.com/janmbuys/DeepDeepParser}.} 
\end{abstract}

\section{Introduction}

An important goal of Natural Language Understanding (NLU) is to parse sentences 
to structured, interpretable meaning representations that can be used 
for query execution, inference and reasoning.
Recently end-to-end models have outperformed traditional pipeline approaches,
predicting syntactic or semantic structure as intermediate steps, 
on NLU tasks such as 
sentiment analysis and semantic relatedness~\cite{LeM14,KirosEa15}, 
question answering~\cite{HermannEa15} 
and textual entailment~\cite{RocktaschelEa15}.
However the linguistic structure used in applications has 
predominantly been shallow, restricted to bilexical dependencies or trees.

In this paper we focus on robust parsing into linguistically deep representations.
The main representation that we use is Minimal Recursion 
Semantics (MRS)~\cite{CopenstakeFMRS95,CopenstakeFPS05}, which serves as
the semantic representation of the English Resource Grammar 
(ERG)~\cite{Flickinger00}.
Existing parsers for full MRS 
(as opposed to bilexical semantic graphs derived from, but simplifying MRS) 
are grammar-based, performing disambiguation with a maximum
entropy model~\cite{ToutanovaMFO05,ZhangOC07};
this approach has high precision but incomplete coverage.

Our main contribution is to develop a fast and robust parser for full MRS-based 
semantic graphs.
We exploit the power of global conditioning enabled by deep learning 
to predict linguistically deep graphs incrementally. 
The model does not have access to the underlying ERG or syntactic structures
from which the MRS analyses were originally derived.
We develop parsers for two graph-based conversions of MRS, 
Elementary Dependency Structure (EDS)~\cite{OepenL06} and Dependency MRS 
(DMRS)~\cite{Copenstake09}, of which the latter is inter-convertible with 
MRS.

Abstract Meaning Representation (AMR)~\cite{BanarescuEa13} is a graph-based
semantic representation that shares the goals of MRS.
Aside from differences in the choice of which linguistic phenomena are 
annotated, MRS is a compositional representation
explicitly coupled with the syntactic structure of the sentence, 
while AMR does not assume compositionality or alignment
with the sentence structure.
Recently a number of AMR parsers have been 
developed~\cite{FlaniganTCDS14,WangXP15,ArtziLZ15,DamonteCS16},
but corpora are still under active development
and low inter-annotator agreement places on upper bound of $83\%$ F1 on 
expected parser performance~\cite{BanarescuEa13}.
We apply our model to AMR parsing by introducing structure 
that is present explicitly in MRS but not in AMR~\cite{BuysB17a}.

Parsers based on RNNs have achieved state-of-the-art
performance for dependency parsing~\cite{DyerBLMS15,KiperwasserG16}
and constituency parsing~\cite{VinyalsEa15,DyerKBS16,CrossH16a}.
One of the main reasons for the prevalence of bilexical dependencies and
tree-based representations is that they can be parsed with 
efficient and well-understood algorithms.
However, one of the key advantages of deep learning is the ability to make 
predictions conditioned on unbounded contexts encoded with RNNs;
this enables us to predict more complex structures without increasing 
algorithmic complexity.
In this paper we show how to perform linguistically deep parsing with RNNs.

Our parser is based on a transition system for semantic graphs.
However, instead of generating arcs over an ordered, fixed set of nodes 
(the words in the sentence), we generate the nodes and their alignments
jointly with the transition actions.
We use a graph-based variant of the arc-eager transition-system.
The sentence is encoded with a bidirectional RNN.
The transition sequence, seen as a graph linearization, can be
predicted with any encoder-decoder model, but we show that 
using hard attention, predicting the alignments with a pointer
network and conditioning explicitly on stack-based features improves
performance.
In order to deal with data sparsity candidate lemmas are predicted as a 
pre-processing step, so that the RNN decoder predicts unlexicalized node
labels.

We evaluate our parser on DMRS, EDS and AMR graphs.
Our model architecture improves performance 
from $79.68\%$ 
to $84.16\%$ F1 over an attention-based encoder-decoder baseline.
Although our parser is less accurate that a high-precision grammar-based
parser on a test set of sentences parsable by that grammar, 
incremental prediction and GPU batch processing 
enables it to parse $529$ tokens per second, against $7$ tokens per second for 
the grammar-based parser.
On AMR parsing our model obtains $60.11\%$ Smatch. 

\section{Meaning Representations}

\begin{figure}
\centering
\includegraphics[scale=0.42]{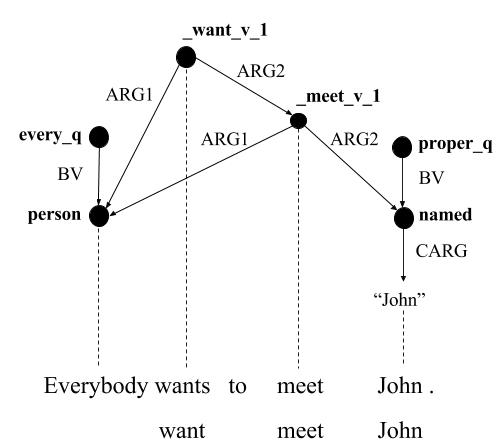}
\caption{Semantic representation of the sentence 
``Everybody wants to meet John.'' 
The graph is based on the Elementary Dependency Structure (EDS) 
representation of Minimal Recursion Semantics (MRS). 
The alignments are given together with the corresponding tokens,
and lemmas of surface predicates and constants.}
\label{fig:eds-graph}
\end{figure}

We define a common framework for semantic graphs in which
we can place both MRS-based graph representations (DMRS and EDS) and AMR.
Sentence meaning is represented with rooted, labelled, 
connected, directed graphs~\cite{KuhlmannO16}.
An example graph is visualized in Figure~\ref{fig:eds-graph}.
representations.
Node labels are referred to as \emph{predicates} (\emph{concepts} in AMR)
and edge labels as \emph{arguments} (AMR \emph{relations}).
In addition \emph{constants}, a special type of node modifiers, are
used to denote the string values of named entities and numbers (including date 
and time expressions).
Every node is aligned to a token or a continuous span of tokens 
in the sentence the graph corresponds to.

Minimal Recursion Semantics (MRS) is a framework for computational semantics
that can be used for parsing or generation~\cite{CopenstakeFPS05}.
Instances and eventualities are represented with logical variables.
Predicates take arguments with labels from a small, fixed set of
roles.
Arguments are either logical variables or handles, designated 
formalism-internal variables.
Handle equality constraints support scope underspecification;
multiple scope-resolved logical representations
can be derived from one MRS structure.
A predicate corresponds to its intrinsic argument and is aligned to a 
character span of the (untokenized) input sentence.
Predicates representing named entities or numbers are parameterized by 
strings.
Quantification is expressed through predicates that bound instance variables,
rather than through logical operators such as $\exists$ or $\forall$.
MRS was designed to be integrated with feature-based grammars such as 
Head-driven Phrase Structure Grammar (HPSG)~\cite{PollardS94} or 
Lexical Functional Grammar (LFG)~\cite{KaplanB82}.
MRS has been implement the English Resource Grammar (ERG)~\cite{Flickinger00},
a broad-coverage high-precision HPSG grammar.

\newcite{OepenL06} proposed Elementary Dependency Structure (EDS),
a conversion of MRS to variable-free dependency 
graphs which drops scope underspecification.
\newcite{Copenstake09} extended this conversion to avoid information loss,
primarily through richer edge labels.
The resulting representation, Dependency MRS (DMRS), can be converted back
to the original MRS, or used directly in MRS-based 
applications~\cite{CopenstakeEa16}.
We are interested in the empirical performance of parsers for both of these
representations: while EDS is more interpretable as an independent semantic 
graph representation, DMRS can be related back to underspecified logical 
forms.
A bilexical simplification of EDS has previously 
been used for semantic dependency parsing~\cite{OepenEa14,OepenEa15}.
Figure~\ref{fig:eds-graph} illustrates an EDS graph.

MRS makes an explicit distinction between surface and abstract
predicates (by convention surface predicates are prefixed by an underscore).
Surface predicates consist of a lemma followed by a coarse 
part-of-speech tag and an optional sense label.
Predicates absent from the ERG lexicon are represented by their 
surface forms and POS tags.
We convert the character-level predicate spans given by MRS to token-level 
spans for
parsing purposes, but the representation does not require gold tokenization.
Surface predicates usually align with the span of the token(s) they represent,
while abstract predicates can span longer segments. 
In full MRS every predicate is annotated with a set of morphosyntactic 
features, encoding for example tense, aspect and number information;
we do not currently model these features.

AMR~\cite{BanarescuEa13} graphs can be represented in the same
framework, despite a number of linguistic differences with MRS.
Some information annotated explicitly in MRS is latent in AMR,
including alignments and the distinction between
surface (lexical) and abstract concepts.
AMR predicates are based on PropBank~\cite{PalmerGK05}, annotated as 
lemmas plus sense labels, but they form only a subset of concepts.
Other concepts are either English words or special keywords, 
corresponding to overt lexemes in some cases but not others.

\section{Incremental Graph Parsing}
\label{sec:parsing}

We parse sentences to their meaning representations by incrementally 
predicting semantic graphs together with their alignments.
Let $\mathbf{e} = e_1, e_2, \ldots, e_I$ be a tokenized
English sentence, $\mathbf{t} = t_1, t_2, \ldots, t_J$ 
a sequential representation of its graph derivation and 
$\mathbf{a} = a_1, a_2, \ldots, a_J$ an alignment sequence consisting
of integers in the range $1, \ldots, I$.
We model the conditional distribution 
$p(\mathbf{t}, \mathbf{a} | \mathbf{e})$ which decomposes as 
\[ 
\prod_{j=1}^{J} p(a_j | \mathbf{(a,t)}_{1:j-1},  \mathbf{e}) 
     p(t_j | \mathbf{a}_{1:j}, \mathbf{t}_{1:j-1}, \mathbf{e}).  
\] 

We also predict the end-of-span alignments as a seperate sequence
$\mathbf{a^{(e)}}$. 

\subsection{Top-down linearization}

\begin{figure}
\begin{verbatim}
:root( <2> _v_1  
  :ARG1( <1> person  
    :BV-of( <1> every_q ) ) 
  :ARG2 <4> _v_1 
    :ARG1*( <1> person
    :ARG2( <5> named_CARG
      :BV-of ( <5> proper_q ) ) )
\end{verbatim}
\caption{A top-down linearization of the EDS graph 
   in Figure~\ref{fig:eds-graph}, using unlexicalized predicates.}
\label{fig:linear-eds}
\end{figure}

We now consider how to linearize the semantic graphs, before defining 
the neural models to parameterize the parser in section~\ref{sec:models}.
The first approach is to linearize a graph as the pre-order traversal of 
its spanning tree, starting at a designated root node 
(see Figure~\ref{fig:linear-eds}).
Variants of this approach have been proposed for neural constituency 
parsing~\cite{VinyalsEa15}, logical form prediction~\cite{DongL16,JiaL16}
and AMR parsing~\cite{BarzdinsG16,PengWGX17}.

In the linearization, labels of edges whose direction are reversed in 
the spanning tree are marked by adding \texttt{-of}. 
Edges not included in the spanning tree, referred to as \emph{reentrancies}, 
are represented with special edges whose dependents are dummy nodes 
pointing back to the original nodes.
Our potentially lossy representation represents these edges by repeating the 
dependent node labels and alignments, which are recovered heuristically.
The alignment does not influence the linearized node ordering.

\subsection{Transition-based parsing}

Figure~\ref{fig:eds-graph} shows that the semantic graphs we work with can also 
be interpreted as dependency graphs, as nodes are aligned to sentence tokens.
Transition-based parsing~\cite{Nivre08} has been used extensively 
to predict dependency graphs incrementally.
We apply a variant of the arc-eager transition system that has
been proposed for graph (as opposed to tree) 
parsing~\cite{SagaeTsujii08,TitovHMM09,GomezN10} to derive a transition-based 
parser for deep semantic graphs.
In dependency parsing the sentence tokens also act as nodes in the graph,
but here we need to generate the nodes incrementally as the transition-system
proceeds, conditioning the generation on the given sentence.
\newcite{DamonteCS16} proposed an arc-eager AMR parser, but 
their transition system is more narrowly restricted to AMR graphs.

\begin{figure*}
\centering  
\begin{tabular}{l|l|l|c}
Action                &    Stack               &    Buffer           & Arc added \\
\hline
init(1, person)         & [ ]                    & (1, 1, person) & -         \\
  sh(1, every\_q)       & [(1, 1, person)]  & (2, 1, every\_q)   & -          \\  
la(BV)                  & [(1, 1, person)]  & (2, 1, every\_q)   & (2, BV, 1) \\ 
  sh(2, \_v\_1)         & [(1, 1, person), (2, 1, every\_q)] & (2, 1, \_v\_1)   & - \\ 
  re                    & [(1, 1, person)] & (3, 2, \_v\_1)  & - \\ 
  la(ARG1)              & [(1, 1, person)] & (3, 2, \_v\_1)  & (3, ARG1, 1)  
\end{tabular}
\caption{Start of the transition sequence for parsing the graph in Figure~\ref{fig:eds-graph}.
The transitions are shift (\texttt{sh}), reduce (\texttt{re}), left arc (\texttt{la}) and 
right arc (\texttt{ra}).
The action taken at each step is given, along with the state of the stack and buffer after 
the action is applied, and any arcs added. 
Shift transitions generate the alignments and predicates 
of the nodes placed on the buffer.
Items on the stack and buffer have the form (\emph{node index, alignment, predicate label}), and 
arcs are of the form (\emph{head index, argument label, dependent index}). 
}
\label{fig:transition-table}
\end{figure*}

The transition system consists of a \emph{stack} of graph nodes being processed 
and a \emph{buffer}, holding a single node at a time.
The main transition actions are \emph{shift}, \emph{reduce}, \emph{left-arc}, 
\emph{right-arc}.
Figure~\ref{fig:transition-table} shows an example transition sequence 
together with the stack and buffer after each step.
The shift transition moves the element on the buffer to the top of the stack,
and generates a predicate and its alignment as the next node on 
the buffer.
Left-arc and right-arc actions add labeled arcs between the buffer and stack top
(for DMRS a transition for undirected arcs is included), but do not change the 
state of the stack or buffer.
Finally, reduce pops the top element from the stack, and predicts its
end-of-span alignment (if included in the representation).
To predict non-planar arcs, we add another transition, which we call
\emph{cross-arc}, which first predicts the stack index of a node
which is not on top of the stack, adding an arc between the head of the buffer 
and that node.
Another special transition designates the buffer node as the root.

To derive an oracle for this transition system, it is necessary to determine
the order in which the nodes are generated.
We consider two approaches.
The first ordering is obtained by performing an in-order traversal of the 
spanning tree, where the node order is determined by the alignment.
In the resulting linearization the only non-planar arcs are reentrancies.
The second approach lets the ordering be monotone (non-decreasing) with respect
to the alignments, while respecting the in-order ordering for 
nodes with the same alignment.
In an arc-eager oracle arcs are added greedily, while a reduce action can
either be performed as soon as the stack top node has been connected to all its
dependents, or delayed until it has to reduce to allow the correct parse
tree to be formed.
In our model the oracle delays reduce, where possible, until the end 
alignment of the stack top node spans the node on the buffer.
As the span end alignments often cover phrases that they head (e.g. for 
quantifiers) this gives a natural interpretation to predicting the span 
end together with the reduce action.

\subsection{Delexicalization and lemma prediction}

Each token in MRS annotations is aligned to at most one surface predicate.
We decompose surface predicate prediction by predicting candidate lemmas for 
input tokens, and delexicalized predicates consisting only of sense labels.
The full surface predicates are then recovered through the predicted alignments.

We extract a dictionary mapping words to lemmas from the ERG lexicon.
Candidate lemmas are predicted using this dictionary, and where no dictionary
entry is available with a lemmatizer.
The same approach is applied to predict constants, along with additional 
normalizations such as mapping numbers to digit strings.

We use the Stanford CoreNLP toolkit~\cite{ManningEa14} to tokenize and lemmatize
sentences, and tag tokens with the Stanford Named Entity Recognizer~\cite{FinkelGM05}.
The tokenization is customized to correspond closely to the ERG tokenization;
hyphens are removed pre-processing step.
For AMR we use automatic alignments and the graph topology to classify 
concepts as surface or abstract. 
The lexicon is restricted to Propbank~\cite{PalmerGK05} predicates;
for other concepts we extract a lexicon from the training data.

\section{Encoder-Decoder Models}
\label{sec:models}

\subsection{Sentence encoder}

The sentence $\mathbf{e}$ is encoded with a bidirectional RNN.
We use a standard LSTM architecture without peephole 
connections~\cite{JozefowiczZS15}.
For every token $e$ we embed its word, POS tag and named entity (NE) tag
as vectors $x_w$, $x_t$ and $x_n$, respectively. 

The embeddings are concatenated and passed through a linear transformation
\[ g(e) = W^{(x)} [x_w; x_t; x_n] + b^{x},
\]
such that $g(e)$ has the same dimension as the LSTM.
Each input position $i$ is represented by a hidden state $h_i$, which is the 
concatenation of its forward and backward LSTM state vectors.

\subsection{Hard attention decoder}

We model the alignment of graph nodes to sentence tokens, $\mathbf{a}$,
as a random variable. 
For the arc-eager model, $a_j$ corresponds to the alignment of the node of the 
buffer after action $t_j$ is executed.
The distribution of $t_j$ is over all transitions and predicates 
(corresponding to shift transitions), predicted with a single softmax.

The parser output is predicted by an RNN decoder.
Let $s_j$ be the decoder hidden state at output position $j$.
We initialize $s_0$ with the final state of the backward encoder.
The alignment is predicted with a pointer network~\cite{VinyalsFJ15}. 

The logits are computed with an MLP scoring the decoder hidden state against
each of the encoder hidden states (for $i = 1, \ldots, I$),
\[
u_j^i = w^T \tanh(W^{(1)} h_i + W^{(2)} s_j). 
\]
The alignment distribution is then estimated by
\[
p(a_j = i | \mathbf{a}_{1:j-1}, \mathbf{t}_{1:j-1}, \mathbf{e}) 
 = \mathrm{softmax}(u_j^i).
\]

To predict the next transition $t_i$, the output vector is conditioned on the 
encoder state vector $h_{a_j}$, corresponding to the alignment:
\begin{align*}
o_j &= W^{(3)} s_j + W^{(4)} h_{a_j} \\
v_j &= R^{(d)} o_j + b^{(d)}, 
\end{align*}
where $R^{(d)}$ and $b^{(d)}$ are the output representation matrix and bias 
vector, respectively.

The transition distribution is then given by
\[ p(t_j | \mathbf{a}_{1:j}, \mathbf{t}_{1:j-1}, \mathbf{e}) = \mathrm{softmax}(v_j). \]

Let $e(t)$ be the embedding of decoder symbol $t$. 
The RNN state at the next time-step is computed as 
\begin{align*}
d_{j+1} &= W^{(5)} e(t_{j}) + W^{(6)} h_{a_j} \\
s_{j+1} &= RNN(d_{j+1}, s_{j}).
\end{align*}

The end-of-span alignment $a_j^{(e)}$ for MRS-based graphs is predicted with another pointer network.
The end alignment of a token is predicted only when a node is reduced from the stack, therefore this
alignment is not observed at each time-step; it is also not fed back into the model.

The hard attention approach, based on supervised alignments, can be contrasted to soft
attention, which learns to attend over the input without supervision.
The attention is computed as with hard attention,
as $\alpha_j^i = \mathrm{softmax}(u_j^i)$. 
However instead of making a hard selection, a weighted average over the encoder vectors is
computed as
$q_j = \sum_{i=1}^{i=I} \alpha_j^i h_i$.
This vector is used instead of $h_{a_j}$ for prediction and feeding to the next time-step.

\subsection{Stack-based model}

We extend the hard attention model to include features based on the transition system stack.
These features are embeddings from the bidirectional RNN encoder,
corresponding to the alignments of the nodes on the buffer and on top of the 
stack.
This approach is similar to the features proposed by \newcite{KiperwasserG16} 
and \newcite{CrossH16} for dependency parsing, although they do not use
RNN decoders.

To implement these features the layer that computes the output vector is 
extended to
\[
o_j = W^{(3)} s_j + W^{(4)} h_{a_j} + W^{(7)} h_{\textrm{st}_0},
\]
where $\texttt{st}_0$ is the sentence alignment index of the element on top of
the stack.
The input layer to the next RNN time-step is similarly extended to
\[
d_{j+1} = W^{(5)} e(t_{j}) + W^{(6)} h_{\textrm{buf}} + W^{(8)} h_{\textrm{st}_0}, \]
where \texttt{buf} is the buffer alignment after $t_j$ is executed.

Our implementation of the stack-based model enables batch processing in static computation graphs,
similar to \newcite{BowmanEa16}. 
We maintain a stack of alignment indexes for each element in the batch, which is
updated inside the computation graph after each parsing action.
This enables minibatch SGD during training as well as efficient batch decoding.

We perform greedy decoding.
For the stack-based model we ensure that if the stack is empty, the next transition
predicted has to be shift.
For the other models we ensure that the output is well-formed during
post-processing by robustly skipping over out-of-place symbols or inserting
missing ones.

\section{Related Work}

Prior work for MRS parsing predominantly predicts structures in the context 
of grammar-based 
parsing, where sentences are parsed to HPSG derivations consistent with the 
grammar, in this case the ERG~\cite{Flickinger00}. 
The nodes in the derivation trees are feature structures, from which
MRS is extracted through unification.
This approach fails to parse sentences for which no valid derivation is found.
Maximum entropy models are used to score the derivations in order to find the 
most likely parse~\cite{ToutanovaMFO05}.
This approach is implemented in the 
PET~\cite{Callmeier00}
and ACE\footnote{\url{http://sweaglesw.org/linguistics/ace/}} parsers.

There have also been some efforts to develop robust MRS parsers.
One proposed approach learns a PCFG grammar to 
approximate the HPSG derivations~\cite{ZhangK11,ZhangEa14}.
MRS is then extracted with robust unification to compose potentially
incompatible feature structures, although that still fails for a small 
proportion of sentences.
The model is trained on a large corpus of Wikipedia text parsed with the 
grammar-based parser. 
\newcite{Ytrestol12} proposed a transition-based approach to HPSG parsing
that produces derivations from which both syntactic and semantic (MRS)
parses can be extracted.
The parser has an option not to be restricted by the ERG.
However, neither of these approaches have results available that can be 
compared directly to our setup, or generally available implementations.

Although AMR parsers produce graphs that are similar in structure to 
MRS-based graphs, most of them make assumptions that are invalid for MRS, 
and rely on extensive external AMR-specific resources.
\newcite{FlaniganTCDS14} proposed a two-stage parser that first predicts
concepts or subgraphs corresponding to sentence segments,
and then parses these concepts into a graph structure.
However MRS has a large proportion of abstract nodes that cannot be 
predicted from short segments, and interact closely with the graph structure.
\newcite{WangXP15,WangXP15a} proposed a custom transition-system for AMR
parsing that converts dependency trees to AMR graphs, relying on assumptions
on the relationship between these.
\newcite{PustHKMM15} proposed a parser based on syntax-based machine 
translation (MT), while AMR has also been integrated into CCG Semantic 
Parsing~\cite{ArtziLZ15,MisraA16}.
Recently \newcite{DamonteCS16} and \newcite{PengWGX17} proposed 
AMR parsers based on neural networks.

\section{Experiments}
\label{sec:experiments}

\subsection{Data}

DeepBank~\cite{FlickingerZK12} is an HPSG and MRS annotation of the
Penn Treebank Wall Street Journal (WSJ) corpus.
It was developed
following an approach known as dynamic treebanking~\cite{OepenFTM04}
that couples treebank annotation with grammar development, in
this case of the ERG.
This approach has been shown to lead to high inter-annotator agreement:
$0.94$ against $0.71$ for AMR~\cite{BenderFOPC15}.
Parses are only provided for sentences for which the ERG has an analysis
acceptable to the annotator -- this means that we cannot evaluate parsing
accuracy for sentences which the ERG cannot parse (approximately $15\%$ of 
the original corpus). 

We use Deepbank version $1.1$, corresponding to ERG 
\texttt{1214}\footnote{\url{http://svn.delph-in.net/erg/tags/1214/}},
following the suggested split of sections $0$ to $19$ as training 
data data, $20$ for development and $21$ for testing.
The gold-annotated training data consists of 35,315 sentences.
We use the LOGON environment\footnote{\url{http://moin.delph-in.net/LogonTop}} 
and the pyDelphin library\footnote{\url{https://github.com/delph-in/pydelphin}}
to extract DMRS and EDS graphs.

For AMR parsing we use LDC2015E86,
the dataset released for the SemEval 2016 AMR parsing 
Shared Task~\cite{May16}.
This data includes newswire, weblog and discussion forum text.
The training set has 16,144 sentences.
We obtain alignments using the rule-based JAMR aligner~\cite{FlaniganTCDS14}.

\subsection{Evaluation}

\newcite{DridanO11} proposed an evaluation metric called
Elementary Dependency Matching (EDM) for MRS-based graphs.
EDM computes the F1-score of tuples of predicates and arguments.
A predicate tuple consists of the label and character span 
of a predicate, while an argument tuple consists of the 
character spans of the head and dependent nodes of the relation, together
with the argument label.
In order to tolerate subtle tokenization differences with respect to 
punctuation, 
we allow span pairs whose ends differ by one character to be matched.

The Smatch metric~\cite{CaiK13}, proposed for evaluating AMR graphs,
also measures graph overlap, but does not rely on sentence alignments
to determine the correspondences between graph nodes. 
Smatch is instead computed by performing inference over graph alignments to 
estimate the maximum F1-score obtainable from a one-to-one matching 
between the predicted and gold graph nodes.

\subsection{Model setup}

Our parser is implemented in 
TensorFlow~\cite{AbadiEa15}.
For training we use Adam~\cite{KingmaB14} with learning rate $0.01$
and batch-size $64$. 
Gradients norms are clipped to $5.0$~\cite{PascanuMB13}. 
We use single-layer LSTMs with dropout of $0.3$ (tuned on the 
development set) on input and output connections. 
We use encoder and decoder embeddings of size $256$, 
and POS and NE tag embeddings of size $32$,
For DMRS and EDS graphs the hidden units size is set to $256$,
for AMR it is $128$.
This configuration, found using grid search and heuristic search
within the range of models that fit into a single GPU,
gave the best performance on the development set under 
multiple graph linearizations.
Encoder word embeddings are initialized (in the first 100 dimensions)
with pre-trained order-sensitive embeddings~\cite{LingDBT15}.
Singletons in the encoder input are replaced with an unknown word
symbol with probability $0.5$ for each iteration.

\subsection{MRS parsing results}

\begin{table}
\centering
\begin{tabular}{l|c|c|c} 
Model           & EDM & EDM$_P$ & EDM$_A$ \\ 
\hline
TD lex   & 81.44 & 85.20 & 76.87  \\
TD unlex & 81.72 & 85.59 & 77.04  \\ 
AE lex   & 81.35 & 85.79 & 76.02  \\ 
AE unlex & 82.56 & 86.76 & 77.54   
\end{tabular}
\caption{DMRS development set results for attention-based encoder-decoder models 
  with alignments encoded in the linearization, for top-down (TD) and arc-eager (AE)
  linearizations, and lexicalized and unlexicalized predicate prediction.}
\label{tab:dmrs-dev-delex}
\end{table}

We compare different linearizations and model architectures
for parsing DMRS on the development data, showing that our approach 
is more accurate than baseline neural approaches. 
We report EDM scores, including scores for predicate
(EDM$_P$) and argument (EDM$_A$) prediction.

First we report results using standard attention-based encoder-decoders,
with the alignments encoded as token strings in the linearization.
(Table~\ref{tab:dmrs-dev-delex}).
We compare the top-down (TD) and arc-eager (AE) linearizations, as well as 
the effect of delexicalizing the predicates (factorizing 
lemmas out of the linearization and predicting them separately.)
In both cases constants are predicted with a dictionary lookup based on the
predicted spans. A special label is predicted for predicates not in the ERG 
lexicon -- the words and POS tags that make up those predicates are recovered 
through the alignments during post-processing.

The arc-eager unlexicalized representation gives the best performance,
even though the model has to learn to model the transition system
stack through the recurrent hidden states without any supervision of the 
transition semantics.
The unlexicalized models are more accurate, mostly due to their ability to 
generalize to sparse or unseen predicates occurring in the lexicon.
For the arc-eager representation, the oracle EDM is $99\%$
for the lexicalized representation and $98.06\%$ for the delexicalized
representation. 
The remaining errors are mostly due to discrepancies between the tokenization 
used by our system and the ERG tokenization. 
The unlexicalized models are also faster to train, as the decoder's output 
vocabulary is much smaller, reducing the expense of computing softmaxes over 
large vocabularies.

\begin{table}
\centering
\begin{tabular}{l|c|c|c} 
  Model           & EDM & EDM$_P$ & EDM$_A$ \\ 
\hline
TD soft  & 81.53 & 85.32 & 76.94 \\
TD hard  & 82.75 & 86.37 & 78.37 \\ 
AE hard  & 84.65 & 87.77 & 80.85  \\ 
AE stack & 85.28 & 88.38 & 81.51 \\ 
\end{tabular}
\caption{DMRS development set results of encoder-decoder models with 
pointer-based alignment prediction, delexicalized predicates and 
hard or soft attention.}
\label{tab:dmrs-dev-point}
\end{table}

Next we consider models with delexicalized linearizations that predict the 
alignments with pointer networks,
contrasting soft and hard attention models (Table~\ref{tab:dmrs-dev-point}).
The results show that the arc-eager models performs better than
those based on top-down representation. 
For the arc-eager model we use hard attention,
due to the natural interpretation of the alignment prediction 
corresponding to the transition system.
The stack-based architecture gives further improvements.

When comparing the effect of different predicate orderings for the 
arc-eager model, we find that 
the monotone ordering performs $0.44$ EDM 
better than the in-order ordering, despite having to parse more non-planar
dependencies.

We also trained models that only predict predicates (in monotone order) 
together with their start spans. 
The hard attention model obtains $91.36\%$ F1 
on predicates together with their start spans with the unlexicalized model,
compared to $88.22\%$ 
for lexicalized predicates and $91.65\%$ for the
full parsing model.

\begin{table}
\centering
\begin{tabular}{l|c|c|c} 
  Model & TD RNN & AE RNN & ACE \\ 
\hline
  EDM       & 79.68 & 84.16 & 89.64 \\
  EDM$_P$   & 83.36 & 87.54 & 92.08 \\ 
  EDM$_A$   & 75.16 & 80.10 & 86.77 \\
  Start EDM & 84.44 & 87.81 & 91.91 \\
  Start EDM$_A$ & 80.93 & 85.61 & 89.28 \\
  Smatch    & 85.28 & 86.69 & 93.50 \\
\end{tabular}
\caption{DMRS parsing test set results, comparing
the standard top-down attention-based and arc-eager stack-based  
RNN models to the grammar-based ACE parser.
} 
\label{tab:dmrs-test}
\end{table}

Table \ref{tab:dmrs-test} reports test set results for various 
evaluation metrics.
Start EDM is calculated by requiring only the start of the alignment
spans to match, not the ends.
We compare the performance of our baseline and stack-based models 
against ACE, the ERG-based parser. 

Despite the promising performance of the model a gap remains between
the accuracy of our parser and ACE.
One reason for this is that the test set sentences will arguably be easier for 
ACE to parse as their choice was restricted by the same grammar that ACE uses.
EDM metrics excluding end-span prediction (Start EDM) show that
our parser has relatively more difficulty in parsing end-span predictions
than the grammar-based parser.

We also evaluate the speed of our model compared with ACE.
For the unbatched version of our model, the stack-based parser parses 
$41.63$ tokens per second, while
the batched implementation parses $529.42$ tokens per second
using a batch size of $128$.
In comparison, the setting of ACE for which we report accuracies parses
$7.47$ tokens per second.
By restricting the memory usage of ACE, which restricts its coverage, we see that ACE 
can parse $11.07$ tokens per second at $87.7\%$ coverage, and $15.11$ tokens per 
second at $77.8\%$ coverage. 

\begin{table}
\centering
\begin{tabular}{l|c|c} 
  Model & AE RNN & ACE \\ 
\hline
  EDM      & 85.48 & 89.58 \\  
  EDM$_P$  & 88.14 & 91.82 \\ 
  EDM$_A$  & 82.20 & 86.92 \\
  Smatch   & 86.50 & 93.52 \\  
\end{tabular}
\caption{EDS parsing test set results.}
\label{tab:eds-test}
\end{table}

Finally we report results for parsing EDS (Table~\ref{tab:eds-test}).
The EDS parsing task is slightly simpler than DMRS, due to the absence of
rich argument labels and additional graph edges that allow the recovery of full MRS.
We see that for ACE the accuracies are very similar, while for our model
EDS parsing is more accurate on the EDM metrics. 
We hypothesize that most of the extra information in DMRS can be obtained
through the ERG, to which ACE has access but our model doesn't. 

An EDS corpus which consists of about $95\%$ of the DeepBank data has also been
released\footnote{\url{http://sdp.delph-in.net/osdp-12.tgz}}, with the goal of 
enabling comparison with other semantic graph parsing formalisms, 
including CCG dependencies and Prague Semantic Dependencies, on the same 
data set~\cite{KuhlmannO16}.
On this corpus our model obtains $85.87$ EDM and $85.49$ Smatch.

\subsection{AMR parsing}

\begin{table}
\centering
\begin{tabular}{l|c|c} 
Model           & Concept F1 & Smatch \\ 
\hline
TD no pointers & 70.16 & 57.95 \\
TD soft       & 71.25 & 59.39  \\ 
TD soft unlex & 72.62 & 59.88  \\   
AE hard unlex & 76.83 & 59.83 \\  
AE stack unlex & 77.93 & 61.21 
\end{tabular}
\caption{Development set results for AMR parsing. All the models except 
the first predict alignments with pointer networks.}
\label{tab:amr-dev}
\end{table}

We apply the same approach to AMR parsing. 
Results on the development set are given in Table~\ref{tab:amr-dev}.
The arc-eager-based models again give better performance, mainly due
to improved concept prediction accuracy.
However, concept prediction remains the most important weakness of the model;
\newcite{DamonteCS16} reports that state-of-the-art AMR parsers score 
$83\%$ on concept prediction.

\begin{table}
\centering
\begin{tabular}{l|c}
  Model    & Smatch  \\ 
\hline
\newcite{FlaniganTCDS14} & 56 \\ 
\hline
\newcite{WangEa16} & 66.54 \\ 
\newcite{DamonteCS16} & 64 \\ 
\hline
\newcite{PengG16}  & 55  \\ 
\newcite{PengWGX17} & 52  \\ 
\newcite{BarzdinsG16} & 43.3 \\ 
\hline
TD no pointers & 56.56 \\ 
AE stack delex & 60.11 
\end{tabular}
\caption{AMR parsing test set results (Smatch F1 scores).
  Published results follow the number of decimals which were reported.}
\label{tab:amr-test}
\end{table}

We report test set results in Table~\ref{tab:amr-test}.
Our best neural model outperforms the baseline JAMR 
parser~\cite{FlaniganTCDS14}, but still lags behind the performance
of state-of-the-art AMR parsers such as CAMR~\cite{WangEa16}
and AMR Eager~\cite{DamonteCS16}.
These models make extensive
use of external resources, including syntactic parsers and semantic role
labellers. 
Our attention-based encoder-decoder model already outperforms
previous sequence-to-sequence AMR parsers~\cite{BarzdinsG16,PengWGX17}, 
and the arc-eager model boosts accuracy further.
Our model also outperforms a Synchronous Hyperedge Replacement Grammar 
model~\cite{PengG16} which is comparable as it does not make extensive use of 
external resources.

\section{Conclusion}

In this paper we advance the state of parsing by employing deep learning 
techniques to parse sentence to linguistically expressive semantic representations 
that have not previously been parsed in an end-to-end fashion.
We presented a robust, wide-coverage parser for MRS that is faster than existing
parsers and amenable to batch processing.
We believe that there are many future avenues to explore to further increase the 
accuracy of such parsers, including different training objectives, more
structured architectures and semi-supervised learning.

\section*{Acknowledgments}

The first author thanks the financial support of the Clarendon Fund and the 
Skye Foundation. 
We thank Stephan Oepen for feedback and help with data preperation,
and members of the Oxford NLP group for valuable discussions.

\bibliographystyle{acl}
\bibliography{references}


\end{document}